\documentclass{article}
\usepackage{spconf,amsmath,graphicx}
\usepackage{ctable}

\usepackage{algorithm}
\usepackage{algpseudocode}

\usepackage{amsfonts} 

\DeclareMathSymbol{\shortminus}{\mathbin}{AMSa}{"39}

\usepackage{mathpazo} 

\usepackage{siunitx,xfp}
\newcommand\ER[1]{\num{\fpeval{trunc(100-#1,2)}}}

\title{LOW-ACTIVITY SUPERVISED CONVOLUTIONAL SPIKING NEURAL NETWORKS APPLIED TO SPEECH COMMANDS RECOGNITION}
\name{Thomas Pellegrini$^{1,}$\sthanks{This work was partially supported by the French ANR agency within the LUDAU project (ANR-18-CE23-0005-01) and the French "Investing for the Future --- PIA3" AI Interdisciplinary Institute ANITI (Grant agreement ANR-19-PI3A-0004). We used HPC resources from CALMIP (Grant 2020-p20022).}, Romain Zimmer$^{1,2}$, Timoth\'ee Masquelier$^2$}
\address{$^1$IRIT, Universit\'e de Toulouse, Toulouse, France\\
	$^2$CERCO UMR 5549, CNRS -- Universit\'e Toulouse 3, Toulouse, France}

\begin{document}
%
\maketitle
\begin{abstract}
Deep Neural Networks (DNNs) are the current state-of-the-art models in many speech related tasks. There is a growing interest, though, for more biologically realistic, hardware friendly and energy efficient models, named Spiking Neural Networks (SNNs). Recently, it has been shown that SNNs can be trained efficiently, in a supervised manner, using backpropagation with a surrogate gradient trick. In this work, we report speech command (SC) recognition experiments using supervised SNNs. We explored the Leaky-Integrate-Fire (LIF) neuron model for this task, and show that a model comprised of stacked dilated convolution spiking layers can reach an error rate very close to standard DNNs on the Google SC v1 dataset: \ER{94.5}\%, while keeping a very sparse spiking activity, below 5\%, thank to a new regularization term. We also show that modeling the leakage of the neuron membrane potential is useful, since the LIF model outperformed its non-leaky model counterpart significantly. 
\end{abstract}
\section{Introduction}
\label{sec:intro}

Deep Neural Networks (DNNs) are the current state-of-the-art models in many speech related tasks. From a computational neuroscience perspective, DNNs can be seen as rate coding based models~\cite{adrian1926impulses}, in the sense that if a neuron is responsive to a given stimulus, then if we augment the stimulus intensity, the neuron output intensity will also increase. Temporal coding based models~\cite{dayan2003theoretical} try to also take into account information carried by the temporal structure of the stimulus. In the case of Spiking Neural Networks (SNNs), spike timing and delays between spikes is important in order to retrieve patterns in the spike sequences given as input to a model. 

There is a growing interest for SNNs applied to speech recognition tasks, from isolated word and phone recognition~\cite{tavanaei2017bio,wu2018biologically,zhang2015digital,bellec2018long},to large-vocabulary automatic speech recognition (ASR) very recently~\cite{10.3389/fnins.2020.00199}. Reasons are that the audio speech signal is particularly suited to event-driven models such as SNNs, SNNs are also more biologically realistic than DNNs, hardware friendly and energy efficient models, if implemented on dedicated energy-efficient neuromorphic chips. Furthermore, it has been shown recently that SNNs can be trained efficiently, in a supervised manner, using backpropagation with a surrogate gradient trick~\cite{neftci2019surrogate}. This new approach allows to train SNNs as one would do for DNNs.

In this work, we propose to use supervised SNNs for speech command (SC) recognition. We explore the Leaky Integrate-and-Fire (LIF) neuron model for this task, and show that convolutional SNNs can reach an accuracy very close to the one obtained with state-of-the-art DNNs, for this task. Our main contributions are the following: i) we propose to use dilated convolution spiking layers, ii) we define a new regularization term to penalize the averaged number of spikes to keep the spiking neuron activity as sparse as possible, iii) we show that the leaky variant of the neuron model outperforms the non-leaky one (NLIF), used in~\cite{10.3389/fnins.2020.00199}.

In order to facilitate reproducibility, our code using PyTorch is available online\footnote{https://github.com/romainzimmer/s2net}.

\section{Overview of the Leaky Integrate-and-Fire model (LIF)}
\label{sec:lif}


The Leaky Integrate-and-Fire (LIF) model~\cite{stein1967some} is a phenomenological model of a biological spiking neuron, that describes how the neuron membrane potential $U$ behaves through time, given an input current $I$ comprised of spikes incoming from afferent neurons. Despite its simplicity, this model is very popular in computational neuroscience, for studies of neural coding, memory,
and network dynamics~\cite{gerstner2002spiking}. From a machine learning point of view, it was shown to outperform more complex models at predicting real neuron activity~\cite{Kobayashi2009,Gerstner2009}, and it is the basic brick of many recent Spiking Neural Networks (SNNs).

In Integrate-and-Fire (IF) models, a spike (output pulse) $\delta(t-t^f)$ --- $\delta$ being the Dirac $\delta$-function --- is generated at the firing time $t^f$ when the potential $U$ crosses from below a threshold value $b$. The potential is then instantly reset to a new value $U_{\mathrm{rest}}$. In the standard leaky variant, the sub-threshold dynamics of the membrane potential of the $i^{th}$ neuron are described by the differential equation Eq.~\ref{eq:lif}~\cite{neftci2019surrogate}:

\begin{equation}
    \tau_{\mathrm{mem}} \frac{d U_i}{d t} = - (U_i(t) - U_{\mathrm{rest}}) + R I_i(t)
    \label{eq:lif}
\end{equation}

where $U_i(t)$ is the membrane potential at time $t$, $U_{rest}$ is the resting membrane potential, $\tau_{\mathrm{mem}}$ is the membrane time constant, $I_i$ is the current injected into the neuron and R is the membrane resistance.  When $U_i$ exceeds a threshold $B_i$, the neuron fires and $U_i$ is decreased. The $-(U_i - U_{\mathrm{rest}})$ term is the leak term that drives the potential towards $U_{\mathrm{rest}}$. 

If there is no leak, the model is called Non-Leaky Integrate and Fire (NLIF) and the corresponding differential equation is:

\begin{equation}
    \tau_{\mathrm{mem}} \frac{d U_i}{d t} = R I_i
    \label{eq:nlif}
\end{equation}

Without loss of generality, we take $R = 1$ hereafter. 

The input current can be defined as the projection of the input spikes along the preferred direction of neuron $i$ given by $W_i$, the $i^{th}$ row of W,  the synaptic weight matrix:

\begin{equation}
    I_i = \sum_j W_{ij} S^{\mathrm{in}}_j 
\end{equation}

where $S_j(t) = \sum_k \delta(t-t_k^j)$ if neuron $j$ fires at time $t= t_1^j,t_2^j,...$. With this formulation, the potential will rise instantaneously when input spikes are received. Alternatively, the current can be governed by a leaky integration of these projections, similar to the membrane potential. For the sake of simplicity, we chose to use the instantaneous formulation for the input current. 


The differential equations of LIF models can be approximated by linear recurrent equations in discrete time. Introducing a reset term $U^R_i[n]$ for the potential, the neuron dynamics can now be fully described by the following equations~\cite{zimmer2019technical}. 

\begin{align} 
    U^R_i[n\shortminus1] & = b_i ||W_i||^2  S_i^{l}[n\shortminus1] \nonumber\\
    I_i[n] & = \sum_j W_{ij} S_j^{l\shortminus1}[n] \nonumber\\ 
    U_i[n] & = \beta (U_i[n\shortminus1]- U^R_i[n\shortminus1]) + I_i[n] \label{eq:pot}\\
    S_i^{l}[n] & = \Theta(U_i[n] - b_i||W_i||^2)\nonumber
\end{align}

where $\beta = \exp(-\frac{\Delta t}{\tau_{\mathrm{mem}}})$, with $\Delta t$ is a time step, $\Theta$ is the Heaviside step function and $b_i$ is the threshold of neuron $i$. 

With these equations, LIF neurons can be modeled as Recurrent Neural Network (RNN) cells whose state and output at time step $n$ are given by $(U[n], I[n])$ and $S[n]$, respectively~\cite{neftci2019surrogate}.

In practice, we used a trainable threshold parameter $b_i$, such that:

\begin{equation}
S_i[n] = \Theta(\frac{U_i[n]}{\| W_i\| ^2 + \epsilon} - b_i)
\end{equation}

with $b_i$ initialized to 1 and $\epsilon = 10^{-8}$.
We normalize the potential with $\| W_i\| ^2$ to avoid squashing the value of the gradients during training. 
Note that we may choose to also optimize the leak parameters $\beta$, during training. Impact of learning those and the thresholds will be discussed in Section~\ref{sec:ablation}.


In~\cite{10.3389/fnins.2020.00199}, the authors based their SNNs on a non-leaky variant of IF models. In this case, solving Eq.~\ref{eq:nlif} and with our notations, Eq.~\ref{eq:pot} boils down to:

\begin{equation}
    U_i[n] = U_i[n\shortminus1]- U^R_i[n] +  I_i[n]\label{eq:nlfi}
\end{equation}

We will report results using this NLIF formulation in Section~\ref{sec:ablation}.

\section{Surrogate Gradient}


In order to train SNNs in a supervised fashion, just as we do in standard deep learning using the back-propagation algorithm and stochastic gradient descent, we need to address the issue regarding the calculation of the gradient of the threshold function. Indeed, the gradient of the Heaviside function is zero everywhere and is not defined at zero. No gradient would be propagated unless we approximate it by a smooth function. That has been proposed in~\cite{neftci2019surrogate}, in the name of a \textit{surrogate gradient} function. In our work, we approximate the gradient of the Heaviside step function by the gradient of a sigmoid function, with a scale parameter $a \geq 0$ to control the quality of the approximation. The sigmoid function we use is: 

\begin{equation}
\text{sig}_a(x)=1/(1+\exp{(-ax)})
\label{eq:sigmoide}
\end{equation}

where $a$ is the scale parameter, $x$ a real-valued input. 
The larger $a$, the steeper the curve. This hyperparameter can be set empirically.

Hence, when using our SNN, on the forward pass the Heaviside function is used, whereas on the backward pass, the gradients are computed using the sigmoid gradient:

\begin{equation}
\Theta'(x) \approx \text{sig}_a'(x)= a \; \text{sig}_a(x) \; \text{sig}_a(-x)
\label{eq:surrogate}
\end{equation}


\section{Spiking layers}

The proposed SNN is a feed-forward model with multiple spiking layers, and a decision output layer, sometimes called a readout layer. The input of a spiking layer is a spike sequence, also called a spike train, except for the first layer, for which the input is a multidimensional real-valued signal (log-FBANK coefficients). Each layer outputs a spike train except for the readout layer that outputs real values that can be seen as a linear combination of spikes, and on which predictions are made.

For fully-connected spiking layers, the input current at each time step would be a weighted sum of the input spikes emitted by the previous layer at the given moment (or a weighted sum of the input signal if it is the first layer). The state and the output of the cells are updated following the equations given in Section 2.

Just as in standard deep learning, we expect convolution filters to be pertinent for our task, given that our input are spectrograms. We thus defined and implemented convolutional spiking layers, as described in Algorithm~\ref{algoconv}. In the 2-d case (time$\times$frequency), the input current at each time step is given by a 2-d convolution between a kernel and the input spike train. The output spike train is computed through a for loop on the time steps, using the equations of the LIF model. As a side note, when processing spike trains as input, convolution in time can be seen as propagation delays of the input spikes. It is well known that such delays enrich the network's dynamics~\cite{Izhikevich2006} and expressivity~\cite{Maass1999}.

\begin{algorithm}
\begin{algorithmic}[1]
\caption{Spiking convolution layer algorithm}\label{algoconv}
\State Inputs: input spike train $S^{\text{in}}$
\State Outputs: output spike train $S^{\text{out}}$
\State Initialize the potential $U$ and $S^{\text{out}}$ to zero tensors
\State Convolve $S^{\text{in}}$ with kernels $w$: $\text{cS} \leftarrow \text{conv2d}(S^{\text{in}}, w)$
\For{n=1,$\ldots$,T} 
\State Compute $R[n]$ 
\State Get input current at time step $n$: $I[n]\leftarrow \text{cS}[n]$ 
\State Compute $U[n]$
\State Apply threshold function to compute $S^{\text{out}}[n]$
\EndFor
\end{algorithmic}
\end{algorithm}




We used the readout layer proposed in~\cite{neftci2019surrogate}, with non-firing neurons (no reset). For classification tasks, the dimension of the output is equal to the number of labels. The label probabilities are given by a Softmax function applied to the maximum value over time of the membrane potential of each neuron.

In practice, we have found that using time-distri\-buted fully connected layer and taking the mean activation of this layer over time as output makes training more stable. Thus, the output is the mean over time of a linear combination of input spikes.

\section{Regularizing the spiking activity}

A desirable property of SNNs would be energy efficiency, meaning that their neuron spiking activity should be as sparse as possible. We would like the number of emitted spikes by each spiking layer as small as possible, while still performing the requested task with high accuracy. This property is also desirable from a biological point of view, since biological neurons are very energy-efficient and emit limited amounts of spikes in a given amounts of time. If sparse, patterns of activity might also be more explainable.

In order to enforce sparse spiking activity, one could apply a $\ell _1$ or $\ell _2$ regularization on the total number of spikes emitted by each layer. However, due to the surrogate gradient, some neurons will be penalized even if they have not emitted any spike. As a regularization term, we propose to use the squared number of spikes at each layer $l$:




$$L_r(l) = \frac{1}{2KN} \sum_n \sum_k S_k^2[n] $$

where $K$ is the number of neurons and $N$ is the number of time steps. Using $S_k[n]^2$ instead of $S_k[n]$ is a simple way to ensure that the regularization will not be applied to neurons that have not emitted any spikes, i.e. for which $S_k[n]=0$. 

Indeed,

$$\frac{d S^2_k[n]}{d U_k[n]} = 2S_k[n] \text{sig}'_{a} (U_k[n])$$

which is zero when $S_k[n]=0$. We will report the regularization impact in the experiment section.

\section{Google Speech Commands dataset and preprocessing}

The Google Speech Commands dataset~\cite{DBLP:journals/corr/abs-1804-03209} is a dataset of short audio recordings (at most 1 second, sampled at 16 kHz) of 30 different commands pronounced by different speakers for its first version and 35 for the second. All experiments were conducted on the first version of the dataset. The task considered is to discriminate among 12 classes: silence, unknown, "yes", "no", "up", "down", "left", "right", "on", "off", "stop", "go".
    
Commands that are not in the ten target classes are labeled as unknown and silence training data are extracted from the background noise files provided with the dataset. We used the standard validation and testing subsets provided with the corpus to evaluate the accuracy of our approach. 

Forty log-Mel coefficients were extracted from raw signals with LibROSA~\cite{brian_mcfee_2015_18369}, using a Mel scale defined between 20 Hz and 4000 Hz. We used a window size of 30 ms and a hop length of 10 ms, which also corresponds to the time step of the simulation $\Delta t$. The resulting input feature maps are of dimension $100\times 40$ in channels, time and frequency. Finally, the spectrograms are re-scaled to ensure that the signal in each frequency band has a variance of 1 across time.

\begin{figure}
    \centering
    \includegraphics[width=0.5\linewidth]{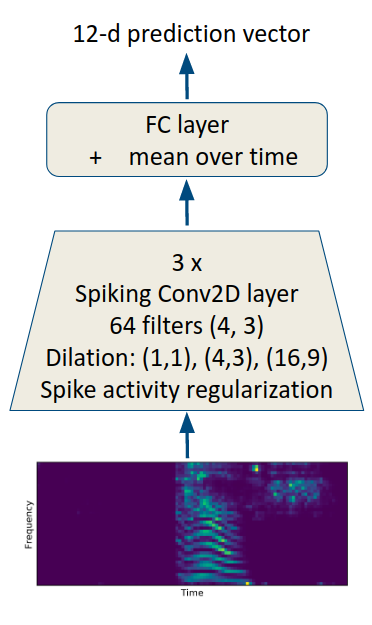}
    \caption{Proposed SNN architecture.}
    \label{fig:model}
\end{figure}

\section{Model architecture}

Fig.~\ref{fig:model} shows the architecture of our proposed SNN for the speech commands recognition task. 

We used three stacked convolution spiking layers and an output/readout layer, which is a time-distributed fully connected layer with twelve output neurons for the twelve classes to be predicted. Each convolution layer has one $\beta$ learnable parameter controlling the time constant of the layer, $C$ channels and one learnable threshold parameter $b$ per channel.

Table \ref{tab:architecture} shows the details of our best model. Kernels are of size $H=4$ along the time axis and $W=3$ along the frequency axis. All convolutional layers have a stride of 1 and dilation factors of $d_H$ and $d_W$ along time and frequency axes respectively. The dilation coefficients were chosen so that the receptive fields $r$ grow exponentially according the layer index: $4\times 3$, $16 \times 9$ and $64\times 27$ for the three convolution layers, respectively. With these settings, the total number of trainable parameters is 0.13M parameters.

The scale $a$ of the surrogate gradient was empirically set to ten. Values below gave worse results and higher values did not bring improvements.

\begin{table}[ht!]
\centering
\begin{tabular}{ |c|c c c c c c| } 
\hline
Conv. layer & $C$ & $H$ & $W$ & $d_H$ & $d_W$ & $r$ \\
\hline
1 & 64 & 4 & 3 & 1 & 1 & $4\times 3$\\
2 & 64 & 4 & 3 & 4 & 3 & $16 \times 9$\\
3 & 64 & 4 & 3 & 16 & 9 & $64\times 27$\\
\hline
\end{tabular}
\caption{Convolutional spiking layer characteristics. $C$: number of channels, $H$, $W$: kernel size in time and frequency, $d_H$, $d_W$: dilation coefficients in time and frequency, $r$: receptive fields.}
\label{tab:architecture}
\end{table}

\begin{figure*}[htb]
\begin{minipage}[b]{.33\linewidth}
  \centering
  \centerline{\includegraphics[width=6cm,height=3.5cm]{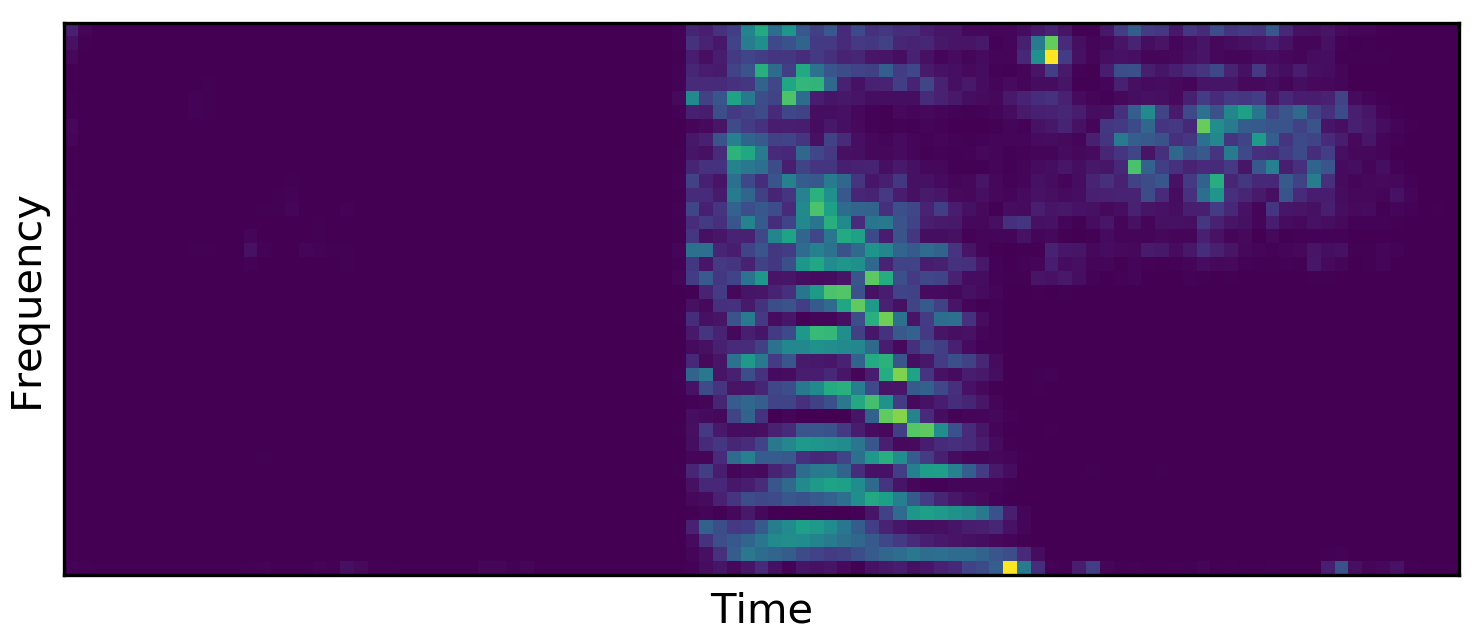}}
  \label{fig:spectro_off}
  \centerline{(a) Log-Mel spectrogram}\medskip
\end{minipage}
\hfill
\begin{minipage}[b]{0.33\linewidth}
  \centering
  \centerline{\includegraphics[width=6cm,height=3.45cm]{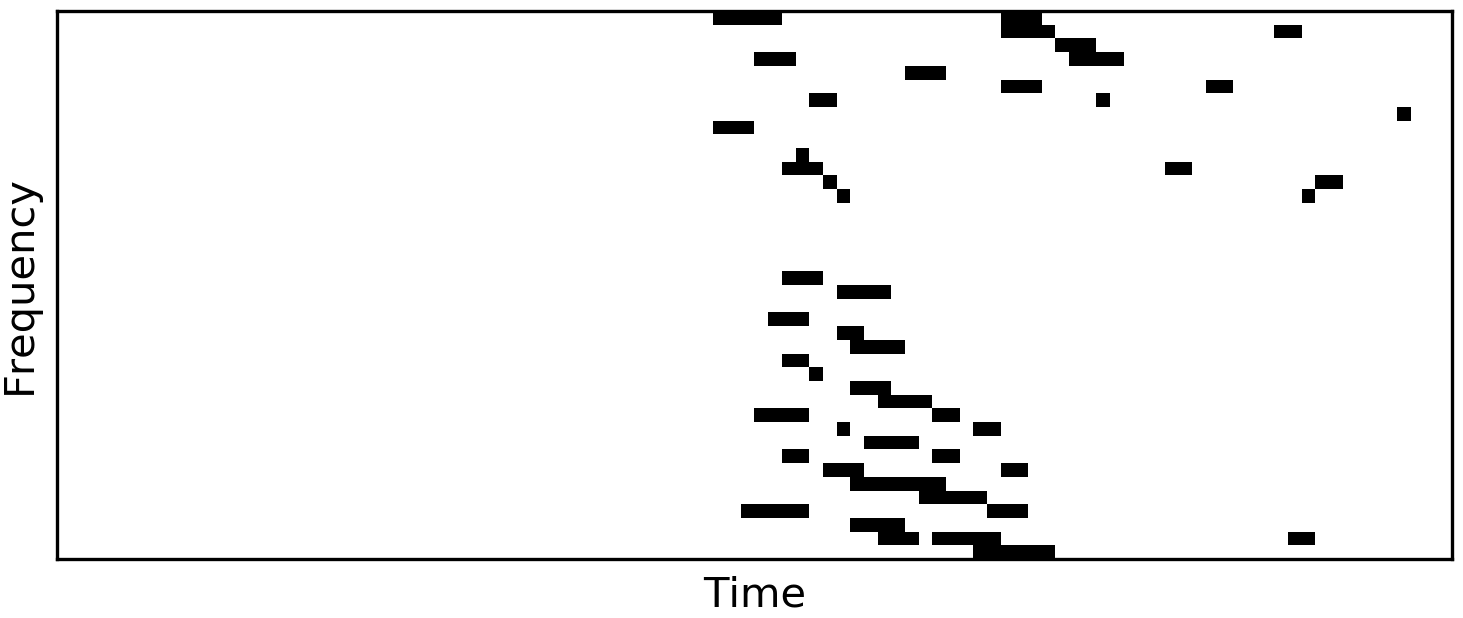}}
\label{fig:spike_train_off}
  \centerline{(b) Spike train, for one channel (first layer)}\medskip
\end{minipage}
\hfill
\begin{minipage}[b]{0.33\linewidth}
  \centering
  \centerline{\includegraphics[width=6cm,height=3.45cm]{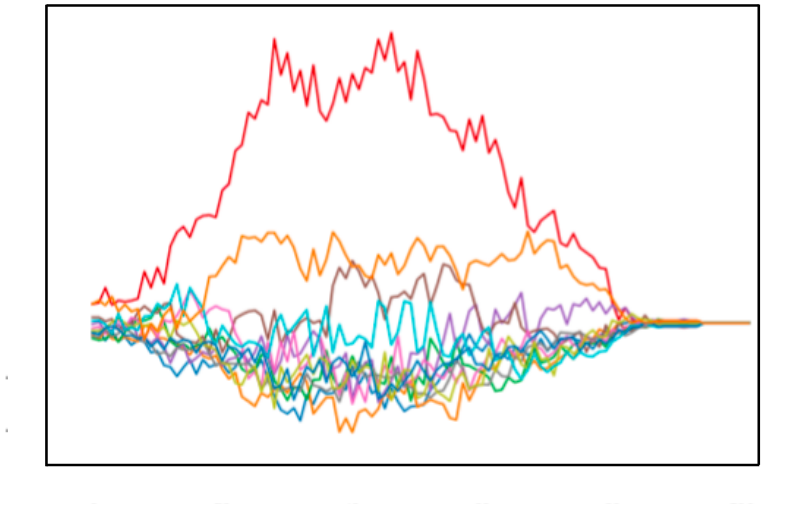}}
\label{fig:output}
  \centerline{(c) Model output}\medskip
\end{minipage}
\caption{Example for the spoken word ``off''. (a) Input features given to the SNN, (b) the corresponding spike train emitted by one channel of the first layer, (c) the model output, X-axis: time, Y-axis: predictions for the twelve classes.\label{fig:off}}
\end{figure*}



\section{Training details}

The model was trained using 128-samples minibatches, with the Rectified-Adam optimizer~\cite{liu2019radam}, with a learning rate of $10^{-3}$, for 20 epochs with one epoch of warm-up, an exponential learning rate scheduler with a 0.85 coefficient, and a weight decay of $10^{-5}$. We used a weighted random sampler to address the unbalanced number of samples of each class. The thresholds $b$ (one per channel) and the $\beta$ (one per layer) parameters were randomly initialized with normal distributions of respective means 1 and 0.7, and 0.01 standard deviations. Gradient values were clipped to $[ -5, 5 ]$, $\beta$ to $[0,1]$ and $b$ to $[0, +\infty[$. For each convolution layer $l = 1,...,L$, the spiking activity regularization loss $L_r(l)$ was added with a 0.1 weighting coefficient. One training epoch took 15 minutes on a 16-GB Tesla V100.

\section{Results}

Fig. 2 shows an example of a log-Mel spectrogram for the word ``off'' (a), the corresponding output after the first convolutional spiking layer (b), for one of its channels, and the network output predictions (c), after training. As can be seen, the first layer binarizes the input spectrogram ; each channel with a different threshold. In Fig. 2 (c), the twelve curves correspond to the twelve non-normalized score values at the different time steps (X-axis). As can be seen, one class gets high scores fast, at the start of the recording, indicating that the convolution layers integrated the relevant information. 

Globally speaking, our SNN achieved \ER{94.5}\% error rate (ER) on the standard test subset. In comparison, the best standard deep learning models reported in the literature achieved error rates from 3\% to 6\%~\cite{journals/corr/abs-1808-08929,DBLP:journals/corr/abs-1804-03209}. In~\cite{journals/corr/abs-1808-08929}, for instance, a ResNet-8 (110k trainable parameters, close to our model size of 131k params), and a ResNet-16 (248k parameters) models reached \ER{94.1}\% and \ER{95.8}\% ER on the same dataset. Regarding the \textit{unknown} class, precision is lower for this class than for the other classes: 80\% compared to 94\% in average for all the classes. Recall is similar to other classes: about 94\%. The \textit{silence} class is perfectly recognized. 

Our implementation keeps track of the number of spikes emitted by each of the convolution layers of the network. On the test subset, the averaged spiking rates are 2.6\%, 4.9\% and 6.1\%, for the first, second and third convolution layers, respectively. Thus, the layer activation is around 5\%, which is sparse as desired. 

After training, we observe that the $\beta$ values decrease in average with the convolution layer index: 0.60, 0.35, 0.23, for the three layers. The smaller the beta, the faster the leak and the more the neurons forget the past and reacts to the incoming spikes. The values of the thresholds $b$ also follow this pattern in average: 0.9541, 0.9434, 0.9284. This is consistent with the previous observation that the deeper the layer is in the network, the lower the thresholds are and the more spikes are emitted.



\begin{table*}[htbp]
\caption{Results and ablation studies. ER: error rate (\%), \# spikes i: averaged percentage of spikes emitted by the $i^{\text{th}}$ convolution layer, during inference on the test subset. NLIF: Non-Leaky Integrate-and-Fire variant~\cite{10.3389/fnins.2020.00199}.}
\begin{center}
\begin{tabular}{lcccccc}
\toprule
 & Model size  & valid ER & test ER & \#spikes 1  & \#spikes 2 & \#spikes 3 \\
 & (Million params) & (\%) & (\%) & (\%) & (\%) & (\%)\\
\midrule
ResNet-8~\cite{journals/corr/abs-1808-08929} & 0.11 & \_ & \hphantom{1}\ER{94.1} & \_ & \_ & \_ \\
\midrule
Full SNN model (ours) & 0.13 & \hphantom{1}\ER{93.4} & \hphantom{1}\ER{94.5} & \hphantom{1}2.6 & \hphantom{1}4.9 & 6.1 \\
\midrule
$\,$ 1-a) w/o dilation & 0.13 & \ER{84.2} & \hphantom{1}\ER{90.9} & \hphantom{1}4.1 & \hphantom{1}3.7 & 6.0 \\
$\,$ 1-b) w/o dilation, large kernels & 4.22 & \hphantom{1}\ER{94.8} & \hphantom{1}\ER{93.6} & \hphantom{1}1.7 & \hphantom{1}1.3 & 4.7 \\ 
\midrule
$\,$ 2) w/o spik. regul. & 0.13 & \hphantom{1}\ER{94.2} & \hphantom{1}\ER{94.3} & 12.5 & 12.5 & 6.4 \\
\midrule
$\,$ 3-a) NLIF, w/ spik. regul. & 0.13 & \ER{83.7} & \ER{87.2} & \hphantom{1}9.0 & \hphantom{1}6.0 & 7.2 \\
$\,$ 3-b) NLIF, w/o spik. regul. & 0.13 & \ER{81.4} & \ER{87.9} & 11.9 & \hphantom{1}7.2 & 8.3 \\
\bottomrule
\end{tabular}
\label{tab:ablation}
\end{center}
\end{table*}




\section{Ablation studies}
\label{sec:ablation}

In order to estimate the impact of different components of our model, we conducted four complementary experiments: 1) when we use non-dilated convolution layers instead of dilated ones, 2) when we remove the spiking activity regularization, 3) when we simplify the LIF model to be the non-leaky Integrate-and-Fire model (NLIF) from~\cite{10.3389/fnins.2020.00199}, 4) when we freeze the leak and threshold parameters: $\beta$ and $b$.

All the model variants used in this section are SNNs. We did another experiment where we removed the spiking components of the convolution and readout layers, thus turning the SNN into a standard convolutional neural network (CNN), comprised of dilated convolution layers and a time-distributed readout layer. This CNN performed much worse, with \ER{47.2}\% ER obtained on the test set, showing that our proposed architecture has been optimized as an SNN, and other design choices should be made to build an efficient CNN. 

Table~\ref{tab:ablation} shows the results of these experiments, in terms of accuracy values obtained on the valid and test subsets, and the average number of spikes for the three convolution layers (\# spikes 1, 2, 3). The first line of the table reports the numbers for our proposed SNN (``full model''), already discussed in the previous section.

\subsection{Dilated convolutions: 1-a,b}

As can be seen in Table~\ref{tab:ablation}, line 1-a, there is a significant increase in ER when using standard instead of dilated convolutions, from \ER{94.5}\% to \ER{90.9}\% ER on the test subset. When removing dilation, the receptive fields of the three layers are notably reduced, compared to the ones listed in~\ref{tab:architecture}. This explains the accuracy reduction, as confirmed by our second experiment, reported in line 1-b, in which we used the model with no dilation but with convolution kernels with the same receptive fields as in our proposed SNN. This larger model recovers the points lost by the 1-a model, with \ER{94.8}\% and \ER{93.6}\% ER on the validation and test sets, respectively. This model is, nevertheless, much larger than our SNN, with 4.22 Million parameters. Dilated convolutions are a way to achieve similar results but with much less parameters. Finally, as a side comment, it is interesting to see that the large model emit much less spikes: below 2\% for the first two layers and 4.7\% for the third one.

\subsection{Spiking activity regularization: 2}

When we remove the penalization on the spiking activity, accuracy remains very close to the full model one (line 2). As expected, the averaged numbers of spikes are larger than for the full model: 12.5\% for the first two layers, and 6.4\% for the third layer. For this last layer, the difference is small (full model: 6.1\%), indicating that the regularization weight, that we chose to be constant to 0.1 for the three layers, was too small for that layer. It could be useful, in future experiments, to use layer-dependent weights, although we do not expect accuracy improvements from tweaking those.

\subsection{Non-leaky neuron model: 3-a,b}

As explained in Section~\ref{sec:lif}, the LIF model can be simplified by removing the potential leak term from the model. The resulting NLFI model has been used in~\cite{10.3389/fnins.2020.00199} for acoustic modeling. In this case, the potential is governed by Eq.~\ref{eq:nlfi}. We replaced Eq.~\ref{eq:pot} by Eq.~\ref{eq:nlfi} in the convolution layer definition of our model, without any other change. Lines 3-a and 3-b lead to the same conclusion, either with or without spike regularization. ER increased significantly compared to using the LIF model (full model): about \ER{87}-\ER{88}\% on the test set. It indicates that the $\beta$ parameters play an important role.

\subsection{Frozen leak coefficients $\beta$ and thresholds $b$}

We did three more experiments, freezing either the $\beta$ values or the threshold $b$ values, or both. We did not report the results in Table~\ref{tab:ablation} since no significant difference has been found compared to the full model ones. For instance, keeping the initial random values for both $\beta$ and $b$ led to \ER{94.3}\% ER. Similar figures were also found for the spiking activity, with a number of spikes that increase with the depth index of the convolution layers. This finding indicates that learning these model parameters is not needed, at least with the initial values we used for them.

\section{Conclusions}

In this work, we explored the LIF neuron model to define dilated convolution spiking layers for a spoken command recognition application. Contrarily to most works using SNNs applied to speech tasks, in which special mechanisms, usually non-trainable, are needed to first encode the speech input features into some type of neural encoding (spike trains) as a first step to then use SNNs (threshold encoding in~\cite{zhang2019mpd}, a specific layer in~\cite{10.3389/fnins.2020.00199}, etc.), our approach is unified in the sense that the first convolution layer applied to real-valued speech features is trainable and shares the same definition and implementation than the ones processing spike trains as input. Our proposed SNN, trained with back-propagation through time with surrogate gradient, achieved results competitive with standard deep convolutional neural networks.   

We defined a regularization term to penalize the averaged number of spikes to keep the spiking neuron activity as sparse as possible, which is a desirable property both from a biological point of view and for a future potential implementation on low-energy dedicated chips.

Finally, we conducted ablation studies in order to estimate the impact of different components of our approach. In particular, an interesting result is that the LIF neuron model outperformed the simpler non-leaky one (NLFI), used in~\cite{10.3389/fnins.2020.00199} for ASR. 

In future work, we will try to confirm these results in acoustic modeling for speech recognition. 

\bibliographystyle{IEEEbib}
\bibliography{strings,refs}

\end{document}